\documentclass[11pt]{article}
\usepackage{nodalida2021}
\usepackage{times}
\usepackage{multicol}
\usepackage{pbox}
\usepackage{url}
\usepackage{hyperref}
\usepackage{appendix}
\usepackage{latexsym}
\usepackage{graphicx}
\usepackage{float}
\usepackage{listings}
\usepackage{makecell}
\aclfinalcopy 
\def\aclpaperid{29} 

\title{Building a Swedish Open-Domain Conversational Language Model}

\author{Tobias Norlund \\
  Chalmers University of Technology \\
  Recorded Future \\
  {\tt tobiasno@chalmers.se} \\\And
  Agnes Stenbom \\
  Royal Institute of Technology \\
  Schibsted Media Group \\
  {\tt astenbom@kth.se} \\}

\date{}

\begin{document}
\maketitle
\begin{abstract}

 We present on-going work of evaluating the, to our knowledge, first large generative language model trained to converse in Swedish, using data from the online discussion forum Flashback.
 We conduct a human evaluation pilot study that indicates the model is often able to respond to conversations in both a human-like and informative manner, on a diverse set of topics.
 While data from online forums can be useful to build conversational systems, we reflect on the negative consequences that incautious application might have, and the need for taking active measures to safeguard against them. 

\end{abstract}

\section{Introduction}

Dialog is an important means through which machines can exhibit intelligence toward humans, which is interesting from a general AI perspective.
But dialog also constitutes a natural interface for humans to interact with technology, which opens up for a breadth of applications involving complex information acquisition, automation of tasks and smart support systems.
A promising direction towards this goal is the development of open domain conversational systems using large neural networks.

Early approaches to neural conversational systems rely on various forms of Recurrent Neural Networks (RNN) trained autoregressively to model the textual sequences \cite{shang2015neural, vinyals2015neural, sordoni2015neural, serban2016building}.
More recently, as large pre-trained Transformer networks have come to dominate progress in NLP in general \cite{devlin2019bert, Radford2018ImprovingLU, radford2019language,  brown2020language, 2020t5}, approaches such as DialoGPT \cite{dialogpt}, Meena \cite{meena} and Blender \cite{blender} have proven the architecture's applicability in open domain dialog systems as well.

However, as the research effort is predominantly put into making progress on English, the importance of making progress in other languages as well has been noted \cite{ruder2020beyondenglish, wali2020machine}.
Each language is its own unique challenge for many reasons, but the difference in availability of resources is a major one, in particular for data-driven methods.
We argue this is also important to keep the public debate on the risks and ethical aspects of large scale language models open to non-English speaking communities.
Toward those ends, we present the first (to our knowledge) attempt to build a large scale open domain dialog system in Swedish based on data from Flashback, one of the largest social discussion forums in Sweden. 
We also present early indicative results on a human evaluation to assess its response generation capabilities across a wide range of topics.

\section{Data and preprocessing}
Flashback\footnote{http://www.flashback.org} is a Swedish online forum that launched in 1996 and has since grown to become one of the country's most popular social medias \cite{svenskarnaochinternet}. 
In the various sub forums, a breadth of topics are openly discussed including computers and programming, economics, politics, sports and science. 
To the general public however, the forum is also widely known for housing an anonymous safe haven for controversial subjects such as prostitution, drugs and conspiracy theories \cite{flashbacklaglost}.
Due to its consistent popularity over the last two decades, it arguably today makes up Sweden's biggest single source of general conversational text.

On Flashback, posts are chronologically organized into threads.
In a single thread, the discussion is centered around a specific topic typically described by a thread title.
Acknowledging the potential for embedding undesired biases, we have initially chosen to use a complete and unfiltered dump of the forum for this study.

The data was tokenized into strings of BPE tokens \cite{sennrich2016neural} using a customly trained vocabulary. 
Due to Flashback's organization of posts into a single linear feed (unlike the tree structure on e.g. Reddit), it is common that users quote the previous post they respond to, to avoid confusion.
As a quote holds important contextual information to a post, we chose to explicitly include this in the way we formatted the threads.
More details of how the data was formatted into strings can be found in Appendix A.

\section{Model}
Following previous works on open-domain dialogue systems \cite{dialogpt, meena}, we trained an auto-regressive language model using a slightly modified Transformer \cite{transformer} decoder as proposed by Radford et. al. \shortcite{radford2019language}. 
That is, for an input sequence of tokens $x_1, ..., x_n$, the language model is trained to maximize the likelihood of the joint probability:

\begin{equation}
    p(x_1, ..., x_n) = p(x_1)\prod_{i=2}^n p(x_i | x_{i-1}, ..., x_1)
\end{equation}
We denote our model \emph{Flashback-GPT}, where GPT is an acronym for Generative Pre-trained Transformer as first coined by Radford et. al. \shortcite{radford2019language}.
The hyper-parameters chosen are similar to those of the largest variant of GPT-2 \cite{radford2019language}, and are detailed in Table \ref{tab:model_hyperparams}.

\begin{table}\centering
\begin{tabular}{|l|l|}
\hline
 Number of layers       & 48 \\
 Dimensionality         & 1600 \\
 Feed-forward dim       & 5400 \\
 Number of heads        & 16 \\
 Number of parameters   & 1.4B \\
 Max context length     & 400 \\
 Batch size             & 512 \\
 Optimizer              & Adam \\
 Vocabulary size        & 52,000 \\
\hline
\end{tabular}
\caption{Model hyperparameters}\label{tab:model_hyperparams}
\end{table}

The model was trained on 16 Nvidia Tesla V100 SXM2 GPUs for 7 days, equivalent to 86,250 gradient updates.
The learning rate was increased linearly for the first 5,000 steps up until $5\mathrm{e}{-5}$, after which it was kept constant.
We used the \texttt{deepspeed} \cite{rajbhandari2020zero} library to optimize memory efficiency across the devices during training.

\section{Evaluation}

Evaluating natural language generation systems is known to be hard. 
Even though it is common to conduct automatic evaluations due to their low cost, a human evaluation often serves as an additional validation of the results.
However, designing a human evaluation to measure a specific quantity is also not trivial since there is always room for interpretation among the human annotators.

Therefore, we present a pilot study where the main aim is merely to get early indications rather than definite results, and to guide the design of bigger future studies.
We design our pilot to measure our quantity of interest: To which extent is the model capable of participating in social discussion forums across a diverse set of topics?

To that end, we seek to measure two quantities: \emph{humanlikeness} and \emph{informativeness}.
As language models can often be inconsistent and show lack of commonsense knowledge, humanlikeness is supposed to answer if there is anything in a response that seems off, suggesting it has not been written by a human.
However, a response can be humanlike but still uninformative.
The notion of "informativeness" is particularly interesting in our setting as forums can be relatively knowledge centric, and uninformative responses such as \emph{I don't know} add little to the discussion.

\subsection{Study design}\label{sec:study_design}

The study was designed as follows. We select a set of $N$ Flashback threads, held out from training, to be used in the study.
For each thread, we only take the first two or three posts to limit the discussion context.
We then, for each thread, swap the last post for an alternative generated by the model.
Along with the originals, we now have $2N$ threads that we present (in shuffled order) to human annotators.
For each thread, we ask two binary questions to measure humanlikeness and informativeness respectively:
\begin{enumerate}
    \item Is there any indication that the last message was not written by a human?
    \item Do you think that the last message adds information to the discussion?
\end{enumerate}

This draws close resemblance to previous evaluations performed on English systems \cite{dialogpt, meena}.
In Zhang et. al. \shortcite{dialogpt}, humans are asked to rank two alternative responses according to \emph{informativeness}, \emph{humanlikeness} and \emph{relevance}.
In Adiwardana et. al. \shortcite{meena}, humans are instead asked the binary questions whether a response "makes sense" and also whether it is "specific", and the average of the two (Sensibleness and Specificity Average - SSA) is found to correlate with humanlikeness.
For simplicity, we chose to directly ask for humanlikeness instead of the SSA proxy questions.
The complete annotator guideline (Swedish) is included in Appendix B for reference.

For the pilot study, we collected a sample of $N=120$ Flashback threads, stratified across 12 of the top level forums.
We then formed two groups of human annotators with three persons in each group.
Each group was presented 60 threads with generated responses, and 60 original, with no overlap.
The threads included were randomly chosen, except for a few criteria that we employed to prevent the annotators from exploiting obvious surface patterns when answering question 1.
\begin{itemize}
    \item As has been noted previously \cite{blender}, beam search decoding strategies have a tendency to generate shorter responses over longer. We decided to only include threads where the last (human written) response is at most 200 characters.
    \item Since the model supports a maximum sequence length of 400 tokens, we exclude threads where the context is longer than 350 tokens, to leave some room for the generated response.
    \item Since the model often fails to generate correct quotes of previous responses, we remove any quotes from the last (human written) response, and force the model not to generate quotes as well.
\end{itemize}

We include a subset of the threads (both with generated and ground truth responses) in Appendix C.

\subsection{Decoding}
The decoding strategy used to generate responses from neural language models is an important part of the system as a whole \cite{blender}. 
While the commonly employed beam search algorithm is optimizing the joint likelihood for the whole generated sequence, its outputs are known to be generic, unspecific and repetitive \cite{holtzman2020curious, see-etal-2019-makes}.
We chose to use a beam sampling strategy, where we at each step, for each beam, sample from the (re-normalized) top 50 predicted vocabulary items.
This struck a good balance between generating short uninformative responses vs longer incoherent ramblings.
We used a beam size of 6.
The model has a tendency to generate responses such as "duplicate thread, locking //mod", which are commonly found on Flashback but are not very interesting for this study. 
We try to circumvent this by banning the generation of certain distinguishing words, such as "mod".
Finally, to avoid repetitions we also prevent the model from generating repetitions of any 3-grams occurring in the context, or in the generated sequence thus far.

\section{Results and Discussion}

\begin{table}
\begin{tabular}{|l|l|l|}
\hline
 \textbf{}   & \textbf{Flashback-GPT}    & \textbf{Human} \\
\hline
 Humanlike & 68\% (48\%) & 95\% (79\%) \\
 Informative & 48\% (52\%)  & 83\% (74\%) \\
 \makecell{Humanlike + \\ informative} & 46\%    & 83\% \\
\hline
\end{tabular}
\caption{Pilot study results. \emph{Humanlike} is the percentage where the majority response to the first question is \emph{no}. \emph{Informative} is the percentage where the majority response to the second question is \emph{yes}. Numbers in parentheses are percentages of the $120$ threads where all three annotators agreed}\label{tab:main_results}
\end{table}

\begin{figure}
\includegraphics[height=4.5cm]{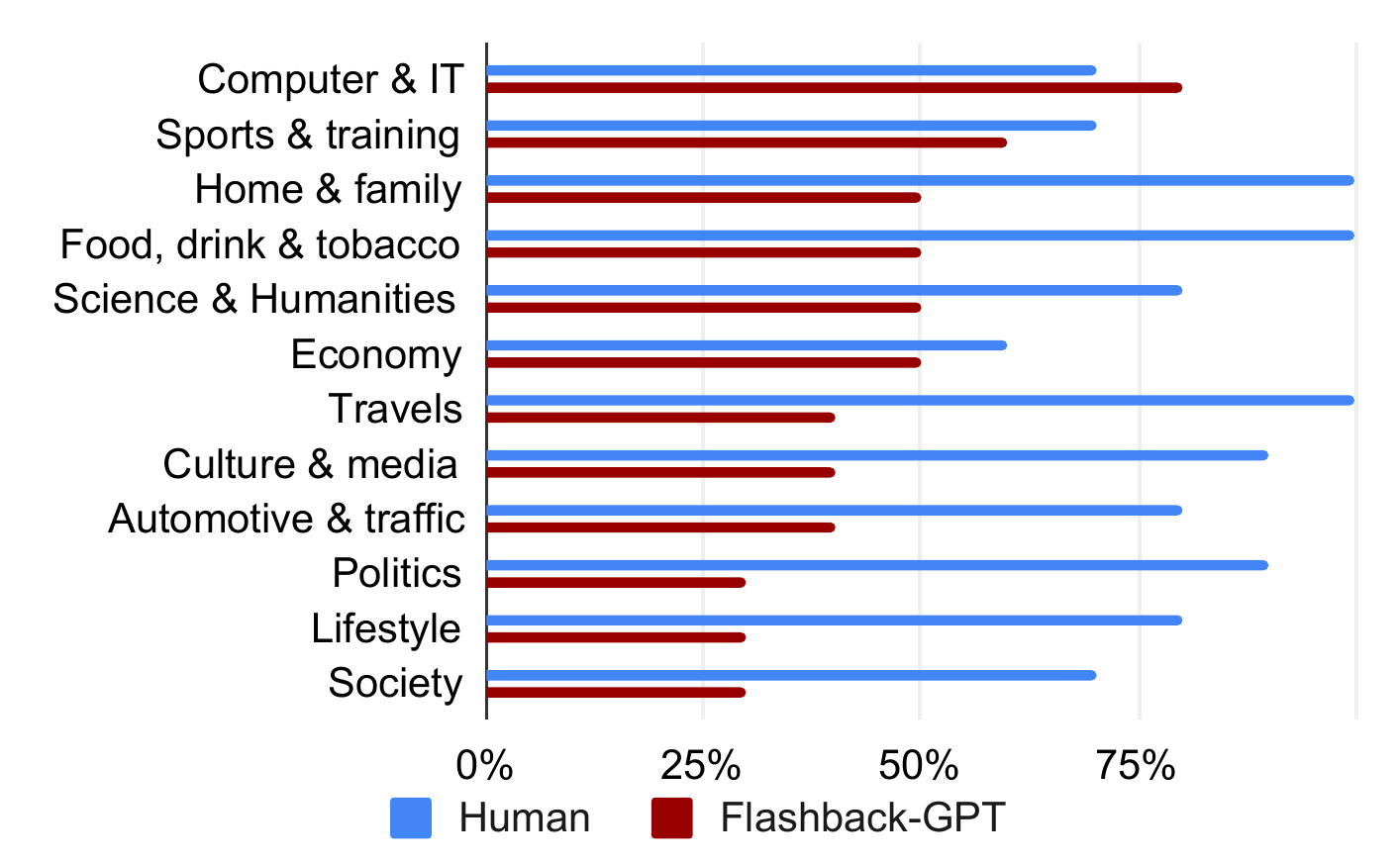}
\caption{Ratio of responses that were deemed both humanlike and informative for each of the evaluated forums}\label{fig:forum_evaluation}
\end{figure}

Results from the study are shown in Table \ref{tab:main_results}.
We judge a thread's humanlikeness and informativeness based on the majority response from the three annotators.
We also report the percentage of threads where all annotators agreed in their responses.

Unsurprisingly, ground truth human responses display a high ratio of humanlikeness, consolidated by a relatively high degree of annotator agreement.
Our model's responses also show signs of humanlikeness, as suggested by the fact that 68\% of its generated responses were deemed plausible to be human-written.
We note however that the annotator agreement is significantly lower compared to ground truth responses, suggesting we could further clarify the humanlikeness question we ask the human annotators.

The model shows less strength on our measure of informativeness, with only 48\% of the model's generated responses were deemed informative to the discussion.
If we compare the amount of threads where the responses were both deemed humanlike and informative, the model's ratio drops to 46\% compared to 83\% for the ground truth responses. While our sample size is too small to draw any statistically significant conclusions, Figure \ref{fig:forum_evaluation} shows the distribution of humanlike + informative responses over their top-level forums.
Interestingly, the top-3 most popular forums (Society, Politics and Culture \& media), which together comprise 41\% of the training data, all perform below average.

Qualitative feedback from the annotators highlight how the model tends to respond with short and straight answers, less prone to vent thoughts and opinions compared to human responses.
Common failure modes include completely misunderstanding the question being asked, or change of topic to a related but irrelevant one. 

Reflecting on the design of the study, we found very few responses were deemed informative but not humanlike (2 of the generated, 0 ground truth).
If the main purpose of a future study is to measure both humanlikeness and informativeness, the question of informativeness might be sufficient.

\section{Broader implications}

Conversational models such as that presented in this paper can be understood as part of a broader transformation of communication. 
As argued by Guzman and Lewis \citeyearpar{Guzman2020}, we are now moving away from the traditional view of communication as anchored in human such.
How we apply and evaluate conversational models going forward may come to alter the way we relate to each other as communicators, and ultimately, humans. 
There is need for informed discussion around what constitutes \textit{desirable} use. 
While highlighting the risks of these emerging technologies could be considered detrimental, we believe it to be an important means towards enabling the inclusion of diverse perspectives in this discussion.

A prominent issue related to NLP is found in the notion of bias. Explicit and implicit biases concerning gender, race or disability can be embedded in e.g. text corpora \citep{Caliskan2017}, word embeddings \citep{bolukbasi2016man} and generative models \citep{Sheng2019}. Employing biased conversational models risks scaling systematic discrimination of various groups in society.

When developing conversational technologies, we must acknowledge that they can be used for malicious purposes. As generative language technology improves and grows in Swedish, so will its ability to manipulate and deceive at scale. As noted by the Swedish Defence Research Agency (FOI), recent developments within generative language technology present risks of increased computer-generated false news and comments – predominately on social media – possibly posing a national security threat \citep{FOI2021}. 

Potential harm must also be considered on the individual level. In 2020, a GPT-3-powered \cite{brown2020language} bot engaged in Reddit-forums with 30 million users about sensitive topics such as suicide and conspiracy theories \citep{MIT}. With the indicative model performance demonstrated in this article, such human-machine communication could soon transpire in Swedish.

\section{Conclusions and Future work}

We demonstrate that Flashback can provide a base on which to build general conversational systems in Swedish.
While our early results suggest the model is often capable to converse across a diverse set of topics, more work remains to examine its utility on various conversational tasks.
We also believe developing methods for grounding the responses in additional data is an interesting direction to further the performance on informativeness in particular.
However, we also believe particular care should be taken as the underlying data is known to contain toxic content.
This points to the importance of putting our model through further scrutiny in following work, to better understand its biases, how they are manifested in downstream tasks, and how they can be mitigated.
Towards those ends, we intend to make the model available for such purposes, and more information is available at \texttt{https://github.com/TobiasNorlund/ \\ flashback-gpt}

\section*{Acknowledgments}
This work was partially supported by the Wallenberg AI, Autonomous Systems and Software Program (WASP) funded by the Knut and Alice Wallenberg Foundation.
The computations were enabled by resources provided by the Swedish National Infrastructure for Computing (SNIC) at Chalmers Centre for Computational Science and Engineering (C3SE) partially funded by the Swedish Research Council through grant agreement no. 2018-05973.

\bibliographystyle{acl_natbib}
\bibliography{nodalida2021}

\begin{thebibliography}{27}
\expandafter\ifx\csname natexlab\endcsname\relax\def\natexlab#1{#1}\fi

\bibitem[{Adiwardana et~al.(2020)Adiwardana, Luong, So, Hall, Fiedel,
  Thoppilan, Yang, Kulshreshtha, Nemade, Lu, and Le}]{meena}
Daniel Adiwardana, Minh-Thang Luong, David~R. So, Jamie Hall, Noah Fiedel,
  Romal Thoppilan, Zi~Yang, Apoorv Kulshreshtha, Gaurav Nemade, Yifeng Lu, and
  Quoc~V. Le. 2020.
\newblock \href {http://arxiv.org/abs/2001.09977} {Towards a human-like
  open-domain chatbot}.

\bibitem[{Bolukbasi et~al.(2016)Bolukbasi, Chang, Zou, Saligrama, and
  Kalai}]{bolukbasi2016man}
Tolga Bolukbasi, Kai-Wei Chang, James Zou, Venkatesh Saligrama, and Adam Kalai.
  2016.
\newblock Man is to computer programmer as woman is to homemaker? debiasing
  word embeddings.
\newblock In \emph{Proceedings of the 30th International Conference on Neural
  Information Processing Systems}, NIPS'16, page 4356–4364, Red Hook, NY,
  USA. Curran Associates Inc.

\bibitem[{Brown et~al.(2020)Brown, Mann, Ryder, Subbiah, Kaplan, Dhariwal,
  Neelakantan, Shyam, Sastry, Askell, Agarwal, Herbert-Voss, Krueger, Henighan,
  Child, Ramesh, Ziegler, Wu, Winter, Hesse, Chen, Sigler, Litwin, Gray, Chess,
  Clark, Berner, McCandlish, Radford, Sutskever, and
  Amodei}]{brown2020language}
Tom~B. Brown, Benjamin Mann, Nick Ryder, Melanie Subbiah, Jared Kaplan,
  Prafulla Dhariwal, Arvind Neelakantan, Pranav Shyam, Girish Sastry, Amanda
  Askell, Sandhini Agarwal, Ariel Herbert-Voss, Gretchen Krueger, Tom Henighan,
  Rewon Child, Aditya Ramesh, Daniel~M. Ziegler, Jeffrey Wu, Clemens Winter,
  Christopher Hesse, Mark Chen, Eric Sigler, Mateusz Litwin, Scott Gray,
  Benjamin Chess, Jack Clark, Christopher Berner, Sam McCandlish, Alec Radford,
  Ilya Sutskever, and Dario Amodei. 2020.
\newblock \href {http://arxiv.org/abs/2005.14165} {Language models are few-shot
  learners}.

\bibitem[{Caliskan et~al.(2017)Caliskan, Bryson, and Narayanan}]{Caliskan2017}
Aylin Caliskan, Joanna~J. Bryson, and Arvind Narayanan. 2017.
\newblock \href {https://doi.org/10.1126/science.aal4230} {Semantics derived
  automatically from language corpora contain human-like biases}.
\newblock \emph{Science}, 356(6334):183--186.

\bibitem[{Devlin et~al.(2019)Devlin, Chang, Lee, and
  Toutanova}]{devlin2019bert}
Jacob Devlin, Ming-Wei Chang, Kenton Lee, and Kristina Toutanova. 2019.
\newblock \href {https://doi.org/10.18653/v1/N19-1423} {{BERT}: Pre-training of
  deep bidirectional transformers for language understanding}.
\newblock In \emph{Proceedings of the 2019 Conference of the North {A}merican
  Chapter of the Association for Computational Linguistics: Human Language
  Technologies, Volume 1 (Long and Short Papers)}, pages 4171--4186,
  Minneapolis, Minnesota. Association for Computational Linguistics.

\bibitem[{Guzman and Lewis(2020)}]{Guzman2020}
Andrea~L Guzman and Seth~C Lewis. 2020.
\newblock \href {https://doi.org/10.1177/1461444819858691} {Artificial
  intelligence and communication: A human–machine communication research
  agenda}.
\newblock \emph{New Media \& Society}, 22(1):70--86.

\bibitem[{Heaven(2020)}]{MIT}
Will~Douglas Heaven. 2020.
\newblock \href
  {https://www.technologyreview.com/2020/10/08/1009845/a-gpt-3-bot-posted-comments-on-reddit-for-a-week-and-no-one-noticed/}
  {A gpt-3 bot posted comments on reddit for a week and no one noticed}.

\bibitem[{Holtzman et~al.(2020)Holtzman, Buys, Du, Forbes, and
  Choi}]{holtzman2020curious}
Ari Holtzman, Jan Buys, Li~Du, Maxwell Forbes, and Yejin Choi. 2020.
\newblock \href {http://arxiv.org/abs/1904.09751} {The curious case of neural
  text degeneration}.

\bibitem[{Internetstiftelsen(2019)}]{svenskarnaochinternet}
Internetstiftelsen. 2019.
\newblock \href
  {https://svenskarnaochinternet.se/rapporter/svenskarna-och-internet-2019/}
  {Svenskarna och internet 2019}.

\bibitem[{Lundén et~al.(2021)Lundén, Melander, Hellquist, Ottosson, Steen,
  and Strindberg}]{FOI2021}
Jenny Lundén, Anders Melander, Elin Hellquist, Björn Ottosson, Liselotte
  Steen, and Anders Strindberg. 2021.
\newblock Strategisk utblick 9 framtida hot.

\bibitem[{Radford(2018)}]{Radford2018ImprovingLU}
Alec Radford. 2018.
\newblock Improving language understanding by generative pre-training.

\bibitem[{Radford et~al.(2019)Radford, Wu, Child, Luan, Amodei, and
  Sutskever}]{radford2019language}
Alec Radford, Jeff Wu, Rewon Child, David Luan, Dario Amodei, and Ilya
  Sutskever. 2019.
\newblock Language models are unsupervised multitask learners.

\bibitem[{Raffel et~al.(2020)Raffel, Shazeer, Roberts, Lee, Narang, Matena,
  Zhou, Li, and Liu}]{2020t5}
Colin Raffel, Noam Shazeer, Adam Roberts, Katherine Lee, Sharan Narang, Michael
  Matena, Yanqi Zhou, Wei Li, and Peter~J. Liu. 2020.
\newblock Exploring the limits of transfer learning with a unified text-to-text
  transformer.
\newblock \emph{Journal of Machine Learning Research}, 21(140):1--67.

\bibitem[{Rasley et~al.(2020)Rasley, Rajbhandari, Ruwase, and
  He}]{rajbhandari2020zero}
Jeff Rasley, Samyam Rajbhandari, Olatunji Ruwase, and Yuxiong He. 2020.
\newblock Deepspeed: System optimizations enable training deep learning models
  with over 100 billion parameters.
\newblock In \emph{Proceedings of the 26th ACM SIGKDD International Conference
  on Knowledge Discovery \& Data Mining}, page 3505–3506, New York, NY, USA.
  Association for Computing Machinery.

\bibitem[{Roller et~al.(2020)Roller, Dinan, Goyal, Ju, Williamson, Liu, Xu,
  Ott, Shuster, Smith, Boureau, and Weston}]{blender}
Stephen Roller, Emily Dinan, Naman Goyal, Da~Ju, Mary Williamson, Yinhan Liu,
  Jing Xu, Myle Ott, Kurt Shuster, Eric~M. Smith, Y-Lan Boureau, and Jason
  Weston. 2020.
\newblock \href {http://arxiv.org/abs/2004.13637} {Recipes for building an
  open-domain chatbot}.

\bibitem[{Ruder(2020)}]{ruder2020beyondenglish}
Sebastian Ruder. 2020.
\newblock {Why You Should Do NLP Beyond English}.
\newblock \url{http://ruder.io/nlp-beyond-english}.

\bibitem[{See et~al.(2019)See, Roller, Kiela, and Weston}]{see-etal-2019-makes}
Abigail See, Stephen Roller, Douwe Kiela, and Jason Weston. 2019.
\newblock \href {https://doi.org/10.18653/v1/N19-1170} {What makes a good
  conversation? how controllable attributes affect human judgments}.
\newblock In \emph{Proceedings of the 2019 Conference of the North {A}merican
  Chapter of the Association for Computational Linguistics: Human Language
  Technologies, Volume 1 (Long and Short Papers)}, pages 1702--1723,
  Minneapolis, Minnesota. Association for Computational Linguistics.

\bibitem[{Sennrich et~al.(2016)Sennrich, Haddow, and
  Birch}]{sennrich2016neural}
Rico Sennrich, Barry Haddow, and Alexandra Birch. 2016.
\newblock \href {https://doi.org/10.18653/v1/P16-1162} {Neural machine
  translation of rare words with subword units}.
\newblock In \emph{Proceedings of the 54th Annual Meeting of the Association
  for Computational Linguistics (Volume 1: Long Papers)}, pages 1715--1725,
  Berlin, Germany. Association for Computational Linguistics.

\bibitem[{Serban et~al.(2016)Serban, Sordoni, Bengio, Courville, and
  Pineau}]{serban2016building}
Iulian~V. Serban, Alessandro Sordoni, Yoshua Bengio, Aaron Courville, and
  Joelle Pineau. 2016.
\newblock Building end-to-end dialogue systems using generative hierarchical
  neural network models.
\newblock In \emph{Proceedings of the Thirtieth AAAI Conference on Artificial
  Intelligence}, AAAI'16, page 3776–3783. AAAI Press.

\bibitem[{Shang et~al.(2015)Shang, Lu, and Li}]{shang2015neural}
Lifeng Shang, Zhengdong Lu, and Hang Li. 2015.
\newblock \href {https://doi.org/10.3115/v1/P15-1152} {Neural responding
  machine for short-text conversation}.
\newblock In \emph{Proceedings of the 53rd Annual Meeting of the Association
  for Computational Linguistics and the 7th International Joint Conference on
  Natural Language Processing (Volume 1: Long Papers)}, pages 1577--1586,
  Beijing, China. Association for Computational Linguistics.

\bibitem[{Sheng et~al.(2019)Sheng, Chang, Natarajan, and Peng}]{Sheng2019}
Emily Sheng, Kai-Wei Chang, Premkumar Natarajan, and Nanyun Peng. 2019.
\newblock \href {https://doi.org/10.18653/v1/D19-1339} {The woman worked as a
  babysitter: On biases in language generation}.
\newblock In \emph{Proceedings of the 2019 Conference on Empirical Methods in
  Natural Language Processing and the 9th International Joint Conference on
  Natural Language Processing (EMNLP-IJCNLP)}, pages 3407--3412, Hong Kong,
  China. Association for Computational Linguistics.

\bibitem[{Sordoni et~al.(2015)Sordoni, Galley, Auli, Brockett, Ji, Mitchell,
  Nie, Gao, and Dolan}]{sordoni2015neural}
Alessandro Sordoni, Michel Galley, Michael Auli, Chris Brockett, Yangfeng Ji,
  Margaret Mitchell, Jian-Yun Nie, Jianfeng Gao, and Bill Dolan. 2015.
\newblock \href {http://arxiv.org/abs/1506.06714} {A neural network approach to
  context-sensitive generation of conversational responses}.

\bibitem[{Vaswani et~al.(2017)Vaswani, Shazeer, Parmar, Uszkoreit, Jones,
  Gomez, Kaiser, and Polosukhin}]{transformer}
Ashish Vaswani, Noam Shazeer, Niki Parmar, Jakob Uszkoreit, Llion Jones,
  Aidan~N Gomez, \L~ukasz Kaiser, and Illia Polosukhin. 2017.
\newblock Attention is all you need.
\newblock In \emph{Advances in Neural Information Processing Systems},
  volume~30, pages 5998--6008. Curran Associates, Inc.

\bibitem[{Vinyals and Le(2015)}]{vinyals2015neural}
Oriol Vinyals and Quoc Le. 2015.
\newblock \href {http://arxiv.org/abs/1506.05869} {A neural conversational
  model}.

\bibitem[{Wali et~al.(2020)Wali, Chen, Mahoney, Middleton, Babaeianjelodar,
  Njie, and Matthews}]{wali2020machine}
Esma Wali, Yan Chen, Christopher Mahoney, Thomas Middleton, Marzieh
  Babaeianjelodar, Mariama Njie, and Jeanna~Neefe Matthews. 2020.
\newblock \href {http://arxiv.org/abs/2007.05872} {Is machine learning speaking
  my language? a critical look at the nlp-pipeline across 8 human languages}.

\bibitem[{Zhang et~al.(2020)Zhang, Sun, Galley, Chen, Brockett, Gao, Gao, Liu,
  and Dolan}]{dialogpt}
Yizhe Zhang, Siqi Sun, Michel Galley, Yen-Chun Chen, Chris Brockett, Xiang Gao,
  Jianfeng Gao, Jingjing Liu, and Bill Dolan. 2020.
\newblock Dialogpt: Large-scale generative pre-training for conversational
  response generation.
\newblock In \emph{ACL, system demonstration}.

\bibitem[{Östman and Aschberg(2015)}]{flashbacklaglost}
Karin Östman and Richard Aschberg. 2015.
\newblock \href
  {https://www.aftonbladet.se/nyheter/a/ddG3Rq/flashback--ett-laglost-land}
  {Flashback – ett laglöst land}.

\end{thebibliography}


\pagebreak
\onecolumn
\begin{appendices}
\gdef\thesection{Appendix \Alph{section}}
\section{Flashback data details}

The data needs to be converted into a textual string format for it to be compatible with a standard language model.
To this end, each thread was formatted into textual \emph{records}.
Listing \ref{lst:record_example} provides an example of a formatted data record used to train the model.
A record can be at most 400 tokens, and as such, threads are often broken up into multiple records.
This means the model will in general not have the full thread context when predicting the next message.

\begin{lstlisting}[caption={Example of a formatted training record. The usernames are anonymized.}, label={lst:record_example}, captionpos=b] 
Dator och IT > Hårdvara: PC
Luft eller vattenkylning till cpu

[user1]:
Jag har lite beslutsångest till vilken kylning jag ska satsa på till min AMD Phenom II X4 965 AM3.
Denna fläkten http://www.komplett.se/k/ki.aspx?sku=456730 eller är det smartare att satsa på vattenkylning?

[user2]:
Citat: [user1]
        Jag har lite beslutsångest till vilken kylning jag ska satsa på till min AMD Phenom II X4 965 AM3.
        Denna fläkten http://www.komplett.se/k/ki.aspx?sku=456730 eller är det smartare att satsa på vattenkylning?
Det där var väl ett jävla åbäk iaf, är du säker på att det inte finns bättre för typ halva priset? Typ Noctua eller liknande?

[user3]:
En vettig fråga är: Vad skall du göra med datorn? Extrem överklockning? Få en tyst dator?
\end{lstlisting}

Table \ref{tab:forum_stats} details the amount of data from each of the top level forums that was used for training.
The dump was collected in September 2020 and in total the data comprised 23.5 GB of raw formatted text.

\begin{table}[!htbp]\centering
\begin{tabular}{|l|l|r|r|}
\hline
\textbf{Top-level forum (swedish)} & \textbf{Top-level forum (english)} & \textbf{Num threads} & \textbf{Num posts} \\ \hline
Samhälle & Society & 230,931 &8,681,841 \\
Politik & Politics & 123,031 &7,578,865 \\
Kultur \& Media & Culture \& Media & 165,929 &6,495,860 \\
Vetenskap \& humaniora & Science \& Humanities & 225,139 &5,130,519 \\
Dator och IT & Computer \& IT & 334,931 &4,833,468 \\
Sport \& träning & Sports \& training & 81,922 &4,475,793 \\
Hem, bostad \& familj & Home \& family & 158,819 &4,055,688 \\
Droger & Drugs & 137,870 &3,551,768 \\
Övrigt & Others & 75,735 &2,164,237 \\
Livsstil & Lifestyle & 81,750 &2,060,600 \\
Sex & Sex & 49,512 &1,335,657 \\
Ekonomi & Economy & 68,078 &1,327,001 \\
Mat, dryck \& tobak &Food, drink \& tobacco & 51,133 &1,286,707 \\
Fordon \& trafik & Automotive \& traffic & 68,078 &1,070,619 \\
Om Flashback & About Flashback & 73,910 &486,536 \\
Resor & Travels & 29,514 &478,150 \\
- & Forum unknown & 181 &71,933 \\
\hline
\textbf{Total} & & \textbf{1,956,463} &\textbf{55,085,242} \\
\hline
\end{tabular}
\caption{Flashback training data statistics}\label{tab:forum_stats}
\end{table}

\pagebreak
\section{Annotation guideline for human evaluation}
\includegraphics[page=1,scale=0.85,angle=-90]{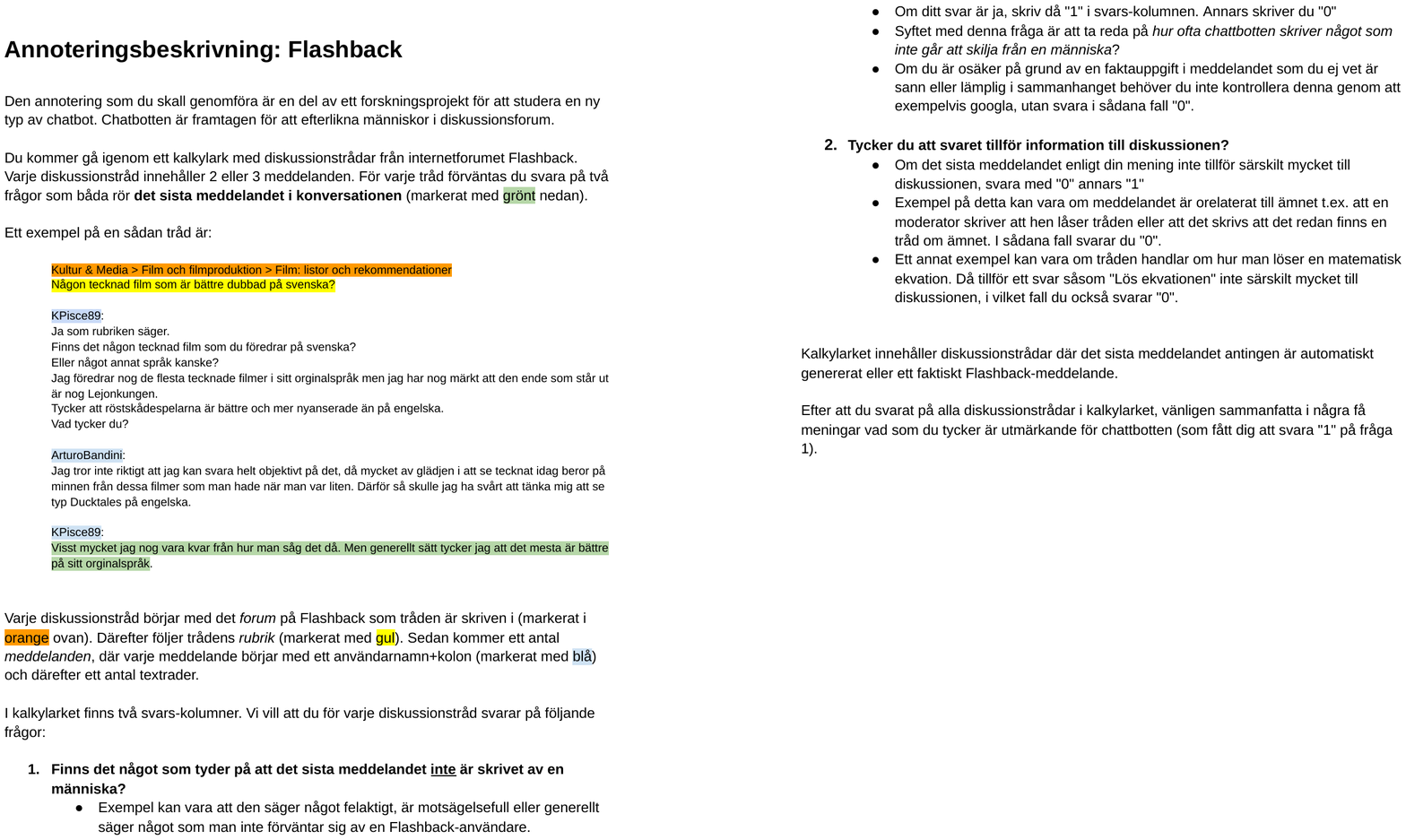}      

\pagebreak
\section{Examples from study}

In the following examples, the last response is generated by the model. Usernames are anonymized.
\begin{lstlisting}
Fordon & trafik > Motorcyklar och mopeder
Off road MC

[user1]:
När jag blir äldre vill jag köra Off road MC, typ Yamaha WR250X verkar nice.
Annars finns det yamaha XT125X. men med mindre klenare motor.
Det är ju bäst att skaffa mc kort med obegränsad motoreffekt, så jag för köra alla typer.
Jag är 175 cm just nu.
1. Hade jag kunnat ha en Yamaha WR250X, så jag inte är för kort?
någon som vet?

[user2]:
när du blir äldre?
e du över 18..?
men nej, tror inte du e för kort att köra off road mc

[user3]:
Vad ska du ha den till?
\end{lstlisting}

\begin{lstlisting}
Vetenskap & humaniora > Fysik, matematik och teknologi > Matematiska och naturvetenskapliga uppgifter
ekvationer som omformas med formler-ma d

[user1]:
5sin4x=3sin2x
lös ekvationen och svara med en decimal?
Jag vet att jag ska flytta över HL i VL. Men sedan vet jag ej vad jag ska göra.

[user2]:
sin(2a) = 2sin(a)cos(a) giver ju i princip svaret.

[user3]:
Du skall multiplicera båda leden med cos(a).
\end{lstlisting}

\begin{lstlisting}
Resor > Övriga resediskussioner
Beställa saker som ligger i planet när jag sätter mig!

[user1]:
Heellu, har en fråga här.. När man beställer saker ifrån tax-free saken så det ligger i sätet när man kommer in i planet, måste man vara 20 och över för vodka o sånt då?
Är inte mer än 19 när jag ska åka, vore gûtt att få med sig en flaska ner dit man nu ska =)

[user2]:
Eftersom du är 19 så misstänker jag att du ska ner till nåt varmt partyställe runt medelhavet, har jag rätt? I så fall är det billigare att köpa den där flaskan på plats och de bryr sig inte om din ålder.

[user3]:
Du behöver inte vara 20 för att köpa sprit i tax-freen.
\end{lstlisting}

\pagebreak
Below are the same examples, but translated to English
\begin{lstlisting}
Automotive & traffic > Motorcycles and mopeds
Off road MC

[user1]:
When I get older I want to drive Off road MC, like Yamaha WR250X seems nice.
Otherwise there is yahama XT125X. but with a weaker engine.
It is best to get the mc license with unlimited power, so I can drive all types.
I'm 175cm right now.
1. Can I have a Yamaha WR250X, or am I too short?
anyone who knows?

[user2]:
when you get older?
are you above 18..?
but no, don't think you're too short to drive off road mc

[user3]:
What are you gonna use it for?
\end{lstlisting}

\begin{lstlisting}
Science & Humanities > physics, mathematics and technology > Mathematical and natural science exercises
reshaping equations with forumlas-ma d

[user1]:
5sin4x=3sin2x
solve the equation and answer with one decimal?
I know I should move right-side over to left-side. But then I don't know what to do.

[user2]:
sin(2a) = 2sin(a)cos(a) basically gives you the answer

[user3]:
You should multiply both sides with cos(a).
\end{lstlisting}

\begin{lstlisting}
Travels > Other travel discussions
Order things to my plane seat

[user1]:
Heellu, got a question here.. When you order stuff from the tax-free thing they lie on your seat when you board the plane, do you have to be 20 or above for vodka and such then?
Won't be more than 19 when I'm going, would be sweet to bring a bottle down to the destination =)

[user2]:
Since you are 19 I'm suspecting you're going down to some warm party place around the Mediterranean, am I right? In such case it is cheaper to buy that bottle in-place and they won't care about your age.

[user3]:
You don't need to be 20 to buy spirits in the tax-free.
\end{lstlisting}

\end{appendices}

\end{document}